\documentclass[12pt]{article}
\usepackage{amsmath}
\usepackage{graphicx}
\usepackage{enumerate}
\usepackage{natbib}
\usepackage{url} % not crucial - just used below for the URL

\usepackage{bibentry}

\usepackage{subcaption}

\usepackage{diagbox}

\usepackage{tikz}
\usetikzlibrary{shapes, arrows}
\usepackage{multirow}
\usepackage{amssymb}
\usepackage{mathtools}
\usepackage{amsthm}
\usepackage{microtype}
\usepackage{graphicx}
\usepackage{booktabs}
\usepackage{float}
\theoremstyle{plain}
\newtheorem{theorem}{Theorem}
\newtheorem{proposition}{Proposition}
\newtheorem{lemma}{Lemma}

\theoremstyle{definition}
\newtheorem{definition}{Definition}

\theoremstyle{remark}

\newtheorem{example}{Example}
\usepackage{bibentry}
\usepackage{color}

\usepackage{mathrsfs}
\usepackage{comment}

\newcommand{\blind}{1}

\begin{document}

\def\spacingset#1{\renewcommand{\baselinestretch}%
{#1}\small\normalsize} \spacingset{1}

%%%%%%%%%%%%%%%%%%%%%%%%%%%%%%%%%%%%%%%%%%%%%%%%%%%%%%%%%%%%%%%%%%%%%%%%%%%%%%

%%%%%%%%%%%%%%%%%%%%%%%%%%%%%%%%%%%%%%%%%%%%%%%%%%%%%%%%%%%%%%%%%%%%%%%%%%%%%%
\if1\blind
{
\title{\bf\Large Learning Optimal Individualized Decision Rules \\with  Conditional Demographic Parity}
\author{Wenhai Cui$^1$, Wen Su$^{2}$, Donglin Zeng$^3$ and Xingqiu Zhao$^1$\\
$^1$Department of Applied Mathematics, The Hong Kong Polytechnic University\\ % wenhai2024.cui@connect.polyu.hk  xingqiu.zhao@polyu.edu.hk
$^2$Department of Biostatistics, City University of Hong Kong\\  %w.su@cityu.edu.hk
$^3$Department of Biostatistics, University of Michigan \\ % dzeng@umich.edu
}

\date{}	
\maketitle
} \fi

\if0\blind
{

  \begin{center}
    {\bf \Large Learning Optimal Individualized Decision Rules \\with Conditional Demographic Parity }
\end{center}
  \bigskip
  \bigskip
  \bigskip
  \medskip
} \fi

\bigskip
\begin{abstract}
Individualized decision rules (IDRs) have become increasingly prevalent in societal applications such as personalized marketing, healthcare, and public policy design. However, a critical ethical concern arises from the potential discriminatory effects of IDRs trained on biased data. These algorithms may disproportionately harm individuals from minority subgroups defined by sensitive attributes like  gender, race, or language.
To address this issue, we propose a novel framework that incorporates demographic parity (DP) and conditional demographic parity (CDP) constraints into the estimation of optimal IDRs. We show that the theoretically optimal IDRs under DP and CDP constraints can be obtained by applying perturbations to the unconstrained optimal IDRs, enabling a computationally efficient solution. Theoretically, we derive convergence rates for both policy value and the fairness constraint term.
The effectiveness of our methods is illustrated through comprehensive simulation studies and an empirical application to the Oregon Health Insurance Experiment.
\end{abstract}

\noindent%
{\it Keywords:} Individual decision rules; Fairness constraints; Demographic parity; Conditional demographic parity; Convergence rate.
\vfill

\newpage
\spacingset{1.9} % DON'T change the spacing!

\section{Introduction}
Individualized Decision Rules (IDRs), which tailor decisions to individuals based on their characteristic, have obtained increasing attention. Applications of IDRs span various domains, including business decision-making \citep{shi2022off}, healthcare \citep{zhao2012estimating}, and social welfare distribution \citep{athey2021policy}.
There are two primary approaches to estimating IDRs. The first is the model-based approach, which includes methods such as Q-learning \citep{dayan1992q, chakraborty2010inference, qian2011performance, song2015penalized} and Advantage-learning (A-learning) \citep{robins2000marginal, murphy2003optimal, robins2004optimal,shi2018high}. Q-learning estimates the conditional expectation of the outcome given treatment and covariates, while A-learning models the contrast of outcomes between treatment and control groups.
The second approach is model-free policy search,  such as Outcome Weighted Learning (OWL) \citep{zhao2012estimating} and its doubly robust version \citep{zhang2012robust, liu2018augmented}. These methods aim to directly maximize the value function over a pre-specified policy class.

With the growing application of IDRs across economic and societal domains, concerns about potential discrimination have become increasingly prominent.
When optimal IDRs are learned from biased or unrepresentative data, automated decision-making algorithms may inadvertently allocate fewer treatment resources to certain groups based on sensitive attributes such as  gender or race. This raises ethical issues, particularly in high-stakes settings such as healthcare  and public policy. For example, in healthcare, older women have been found to be less likely to be admitted to intensive care units and more likely to die following critical illness compared to men \citep{fowler2007sex, penner2010aversive, williams2015racial}.

There are four primary sources of unfairness in IDRs.
First, potential treatment outcomes may be biased due to the use of subjective or discriminatory metrics. For instance, scores assigned by human evaluators, who may have implicit biases against certain subgroups—such as racial minorities or non-native speakers, are used as ground truth to estimate treatment effects.
Second, disparities in treatment quality caused by bias can directly impact outcomes. For example, research has shown that implicit discrimination is significantly associated with adverse birth outcomes \citep{larrabee2021association}. In the U.S., the pregnancy-related mortality ratio between 2011 and 2016 for Black women was more than three times higher than that for White women \citep{genevieve2020structural, hall2015implicit}.
Third, the presence of correlations between sensitive attributes and non-sensitive covariates can indirectly introduce discrimination into the IDRs, even if sensitive attributes are excluded from the modeling process.
Finally, the under-representation of minority subgroups in the data can lead to unreliable estimation of treatment effects for those groups. Small sample sizes can result in high variance and biased estimators.

In view of  such unfairness, demographic parity, which requires that IDRs be statistically independent of sensitive attributes, can be used to reduce  discrimination caused by sensitive attributes and ensure the IDRs align between subgroups. Additionally, in welfare-related decision-making, maximizing expected outcomes is not the sole objective \citep{heidari2018fairness}.
Demographic parity is both an ethical obligation and a legal requirement such as the Civil Rights Act of 1964 and the Equal Credit Opportunity Act of 1974.

Although demographic parity (DP) has been extensively studied in regression and classification tasks, its application within the realm of IDRs remains relatively underexplored.
\cite{kim2023fair} focused on ensuring that the conditional average treatment effect (CATE) satisfies DP, while \cite{wang2025advancing} extended this idea to censored rewards. However, requiring CATE to satisfy DP is stricter than enforcing DP on IDRs.
{First, as treatment outcomes are influenced by various latent factors, imposing DP on the CATE level may lead to significant value loss.
Second, from a decision-making perspective, it is sufficient to focus on the IDR,  as it directly targets optimal treatment assignment without requiring accurate CATE estimation.}
As demonstrated in  example 2 in Section 2, imposing DP on CATE excludes a class of optimal IDRs that satisfy DP-IDRs as defined in our framework, leading to a reduction in policy value.   {\cite{GITR}  proposed a method for  learning a data representation that satisfies DP, and subsequently using this representation as input  to estimate an optimal IDR, referred to as the action fairness IDR (AF-IDR). However, the AF-IDR framework  does not allow policymakers to control  the level of DP. }

Enforcing DP within IDRs using conventional  approaches still remains significant challenges. In-processing methods in machine learning, which typically incorporate proxy constraints for DP into the training objective, offer a relaxed formulation of the fairness criterion and thus cannot strictly guarantee DP \citep{zafar2019fairness,agarwal2019fair}. Moreover, incorporating DP constraints directly into the IDRs framework introduces computational difficulties. The discrete treatment space results in non-smooth constraint formulations, which complicate the optimization process.  In regression or classification tasks,
pre-processing methods attempt to mitigate bias by  injecting noise into the data prior to model training \citep{GITR,wang2019repairing}, while post-processing methods adjust model outputs after training to achieve DP \citep{gordaliza2019obtaining}. However, both approaches inevitably discard valuable information embedded in the original data, resulting in an unnecessary reduction in policy value when such methods are directly applied  to IDRs.  Although some risk-control methods have been proposed for   IDRs \citep{risk2018,liu2024controlling}, they are not directly applicable to fairness-constrained IDRs due to fundamental differences in the formulations of the constraints.

To address these challenges, we  propose a novel framework that estimates  optimal IDRs while directly incorporating a conditional demographic parity (CDP) constraint, which requires that IDRs be conditionally  independent of sensitive attributes given a ``legitimate'' feature.  The DP-IDR is a special case of the CDP-IDR when the “legitimate” feature is constant.
We  establish a  connection between  solving the CDP-IDR and finding the root of a one-dimensional Lagrangian function, and then derive a
 closed-form theoretical solution for the CDP-IDR.
Our work makes five key contributions:
\begin{itemize}
\item[(i)] {\it Direct Enforcement of Demographic Parity:} To the best of our knowledge, this is the first study to directly enforce demographic parity in individualized decision rules  without relaxing the fairness constraint, thereby providing exact fairness guarantees.

\item[(ii)] {\it Incorporation of Conditional Demographic Parity:} This framework is the first to incorporate the conditional demographic parity constraint into IDRs. This formulation eliminates the influence of sensitive attributes while preserving policy value through legitimate, non-discriminatory features that may be correlated with those attributes.

\item[(iii)] {\it Computational Efficiency:} The procedure is computationally efficient. We demonstrate that enforcing CDP is equivalent to adding a fairness-aware perturbation to the unconstrained optimal IDR, thus avoiding the complexities of solving a non-smooth constrained optimization problem.

\item[(iv)] {\it Flexible Trade-off Between Value and Fairness:} The proposed method allows for a flexible trade-off between value optimization and fairness constraints  via a user-specified tolerance  level.

\item[(v)] {\it Estimation Using Deep Neural Networks:} We estimate the optimal IDRs using deep neural networks (DNNs) and establish asymptotic guarantees for both value loss and fairness constraints. Our theoretical analysis shows that the value loss primarily depends on the degree of unfairness, and the proposed estimator asymptotically satisfies the fairness constraints.
\end{itemize}

The remainder of this paper is organized as follows. Section 2 introduces the preliminaries and formal definitions of DP-IDR and CDP-IDR. Section 3 presents the proposed methodology for learning CDP-IDR. Section 4 discusses the estimation procedures, and Section 5 provides theoretical guarantees. Section 6 presents simulation results, while Section 7 applies the method to the Oregon Health Insurance Experiment. Finally, Section 8 concludes with a discussion of the findings and future research.

\section{Preliminaries}

\subsection{Individualized Decision Rules}
Consider a single stage IDR problem. Let
 $\boldsymbol{X} \in \mathcal{X}$ denote a $p$-dimensional covariate vector, and $S \in \mathcal{S}:= \{0,1\}$ represent a binary sensitive attribute (e.g.,  gender, language, or race). {Moreover, we   observe  a categorical variable $L \in \mathcal{L}$ (e.g., credit rating in loan approval settings or poverty level in welfare programs), which  is  regarded as a  ``legitimate''  feature for  decision-making processes.  Let
 $A \in \mathcal{A}:= \{1, -1\}$ be the treatment and $R$ denote the observed outcome. We define $R(1)$ and $R(-1)$  as the potential outcomes under treatments $1$ and $-1$, respectively.} The observed data consist of   $(\boldsymbol{X}, S, L, A, R)$. For simplicity,  we denote $\boldsymbol{Z}:=(\boldsymbol{X}, S, L)$.

An individualized decision rule, denoted by $\mathcal{D}(\boldsymbol{Z})$, is a function mapping from covariate space $\mathcal{Z}$ to the treatment space $\mathcal{A}$. Given an IDR $\mathcal{D}$, the corresponding potential outcome is $R(\mathcal{D}):=R(1)I(\mathcal{D}(\boldsymbol{Z})=1)+R(-1)I(\mathcal{D}(\boldsymbol{Z})=-1)$ and the value function is defined by
$
\mathcal{V}(\mathcal{D}):=E\left[R(\mathcal{D})\right].
$
An optimal IDR, denoted $\mathcal{D}^*$, is the rule that maximizes the value function, i.e.,
\begin{equation}
\begin{aligned}
\mathcal{D}^* = \underset{\mathcal{D}}{\arg \max }
   \mathcal{V}(\mathcal{D}).
   \nonumber
\end{aligned}
\end{equation}

Suppose that $\delta_R(\boldsymbol{Z})=E[ R \mid \boldsymbol{Z}, A=1]-E[R \mid \boldsymbol{Z}, A=-1]$ is the conditional average treatment effect.
To estimate the value via observed data, we assume:
(i) Stable unit treatment value assumption: The observed outcome $R$ equals the potential outcome given treatment, i.e., $R = R(a)$ when $A = a$ for $a \in \mathcal{A}$;
(ii) No unmeasured confounders assumption: The potential outcomes ${R(-1), R(1)}$ are independent of the treatment  $A$ given the covariates $\boldsymbol{Z}$ \citep{rosenbaum1983central, robins2000marginal, robins2004optimal}.
Under these assumptions, we can derive the following equation (see Supplementary A for details):
\begin{equation}
\begin{aligned}
\label{e1_vallue}
2\mathcal{V}(\mathcal{D})-E\left[ E\left\{R \mid \boldsymbol{Z}, A=1\right\}+E\left\{ R \mid \boldsymbol{Z}, A=-1\right\}\right]=E\left\{\delta_R(\boldsymbol{Z})\mathcal{D}(\boldsymbol{Z})\right\}.
\end{aligned}
\end{equation}
Let $\mathcal{D}(\boldsymbol{Z})=2I\{f(\boldsymbol{Z})>0\}-1$, where $f$ is a decision function.
As a result, estimating an  optimal IDR is equivalent to solving:
\begin{equation}
\label{e1}
\begin{aligned}
{f^*}=\underset{f}{\arg \max }
   E \left\{I(f(\boldsymbol{Z})>0)\delta_R(\boldsymbol{Z})\right\},
\end{aligned}
\end{equation}
and the corresponding optimal IDR is $\mathcal{D^{*}}(\boldsymbol{Z})=2I(f^*(\boldsymbol{Z})>0)-1$.

\subsection{Fairness in Individualized Decision Rules}
\begin{figure}[h]
    \centering
    \begin{minipage}{0.45\linewidth}
        \centering
        \begin{tikzpicture}[node distance=2cm, auto]
            \tikzset{
                block/.style = {draw,circle, minimum width=0.5cm, minimum height=0.5cm},
                arrow/.style = {->, >=stealth', shorten >=1pt, shorten <=1pt}
            }
            \node[block] (A) {A};
            \node[block, above of=A] (X) {$\boldsymbol{X}$};
            \node[block, right of=A] (R) {$R$};
            \node[block, above of=R] (S) {$S$};
            \draw[arrow] (A) -- (R);
            \draw[arrow] (X) -- (R);
            \draw[dashed, arrow] (S) -- (R);
             \draw[dashed, arrow] (S) -- (A);
              \draw[arrow] (X) -- (A);
            \draw[dashed, arrow] (X) -- (S);
            \draw[dashed, arrow] (S) -- (X);
        \end{tikzpicture}
    \label{fig2}
    \end{minipage}
    \hspace{0.5cm}
    \begin{minipage}{0.45\linewidth}
        \centering
        \begin{tikzpicture}[node distance=2cm, auto]
            \tikzset{
                block/.style = {draw,circle, minimum width=0.5cm, minimum height=0.5cm},
                arrow/.style = {->, >=stealth', shorten >=1pt, shorten <=1pt}
            }
          \node[block] (A) {$A$};
            \node[block, above of=A] (X) {$\boldsymbol{X}$};
            \node[block, right of=A] (R) {$R$};
            \node[block, right of=X] (E) {$L$};
            \node[block, above of=E] (S) {$S$};

            \draw[arrow] (A) -- (R);
            \draw[arrow] (X) -- (R);
            \draw[dashed, arrow] (S) -- (E);
             \draw[arrow] (E) -- (R);
             \draw[dashed, arrow] (S)to [out=180,in=180](A);
              \draw[arrow] (X) -- (A);
              \draw[arrow] (E) -- (A);
             \draw[dashed, arrow] (X) -- (E);
              \draw[dashed, arrow] (E) -- (X);
             \draw[dashed, arrow] (E) -- (S);
              \draw[dashed, arrow] (E) -- (S);
            \draw[dashed, arrow] (X) -- (S);
            \draw[dashed, arrow] (S) -- (X);
             %\draw[arrow, bend left] (S) to (R);
            \draw[dashed, arrow] (S) to [out=0,in=360] node[midway, right] {} (R);
            %\draw[arrow] (S) -- (Y);
            %\draw[dashed, arrow] (X) -- (S);
           % \draw[dashed, arrow] (S) -- (X);
        \end{tikzpicture}

        \label{fig3}
    \end{minipage}
    \caption{Biased causal diagram.}
\end{figure}
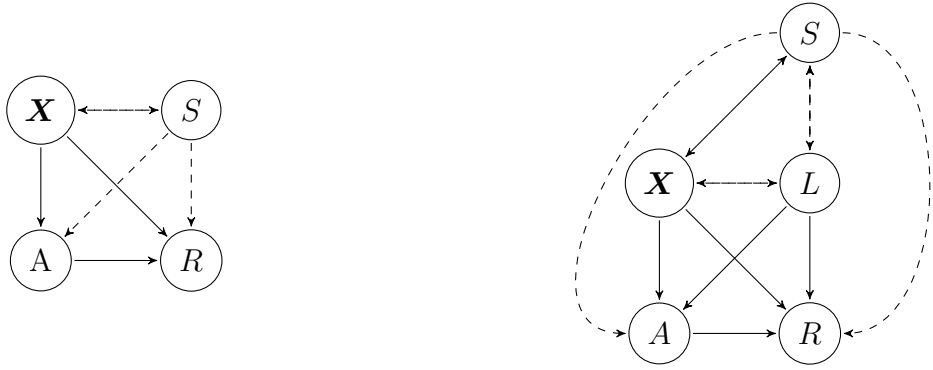

{
To illustrate our motivation, consider the following example:
\begin{example}
Consider a welfare distribution program in which certain applicants are selected to receive free medical care $(A=1)$. The objective of the program is to enhance individuals' physical well-being, measured by a health score $R$ assessed by a physician. Under an unbiased setting, the outcome follows the model:
\begin{equation}
\begin{aligned}
R=(X_1-3)A+\varepsilon,
\nonumber
\end{aligned}
\end{equation}
where $X_1 \sim \mathcal{N}(0,1)$ and $\varepsilon$ is independent noise. The optimal IDR in this case is $ 2I(X_1 > 3)-1$.
Suppose that a biased physician systematically assigns lower scores to individuals from a minority group. Define $S \sim \text{Bernoulli}(p)$ as a sensitive attribute, with $S=1$ indicating minority status and $S$ being independent of $X_1$. Under the biased scenario, the observed data follow the model:
\begin{equation}
R = (X_1 - S - 3)A + \varepsilon,
\nonumber
\end{equation}
leading to the optimal IDR $ 2I(X_1 > 3 + S)-1$. This rule unfairly restricts access to care for the minority group.
\end{example}
In this example, learning an optimal IDR from biased data (see  Figure 1) fails to retrieve the IDR optimal in an unbiased environment, posing a fundamental challenge for fair policy learning.
Furthermore, in practical domains like job training programs (Lyons et al., 2017) or free healthcare allocation (Finkelstein et al., 2012), treatment resources are limited. Therefore, IDRs must not only maximize expected outcomes but also adhere to anti-discrimination laws and societal norms to ensure practical decision-making.}

We begin with formally defining the DP-IDR within the framework of IDRs as follows:
\begin{definition}
(DP-IDR). An IDR $\mathcal{D}: \mathcal{Z}\rightarrow \mathcal{A}$ attains demographic parity, if  it satisfies \begin{equation}
\label{e2}
\begin{aligned}
{P}[\mathcal{D}(\boldsymbol{Z})=a \mid {S}={s}]={P}[\mathcal{D}(\boldsymbol{Z})=a\mid {S}={s}^{\prime}],
\nonumber
\end{aligned}
\end{equation}
for any $ {s}, {s}^{\prime} \in \mathcal{S}$ and $a \in \{-1,1\}$.
\end{definition}
This definition requires that the conditional distribution of DP-IDR be invariant across groups defined by the sensitive attribute $S$, thereby implying statistical independence between the DP-IDR and $S$.

To illustrate the distinction between DP-IDRs and  DP-CATE, which requires the CATE to be independent of sensitive attributes, we provide a simple toy example.
\begin{example}
Suppose that reward is generated from the model:
    $$R=\left\{ (S+1)\operatorname{sign}(X_1-X_2)+S\right\}A+\epsilon,$$ where  $X_1$ and $X_2$ are
independently drawn from a normal distribution $\mathcal{N}(0, 1)$, $S\sim\text{Bernoulli }(0.5)$ is independent of $(X_1, X_2)$  and  $\epsilon \sim \mathcal{N}(0,1)$.
\end{example}
Under this setup, the CATE  is given by $\delta_R(\boldsymbol{Z})= 2(S+1)\operatorname{sign}(X_1-X_2)+2S$, which depends on the sensitive attribute $S$, thereby violating demographic parity for CATE. However, the optimal IDR $\mathcal{D}(\boldsymbol{Z})=2I(X_1-X_2>0)-1$ satisfies DP.

Next, we formally introduce CDP as another fairness criterion for IDRs. CDP requires that the policy satisfy DP within each group defined by a ``legitimate''  variable $L=l$, which CDP is  regarded as consistent with the principles of the European Union Non-Discrimination Law \cite{wachter2020bias}.
\begin{definition}
(CDP-IDR). An IDR $\mathcal{D}: \mathcal{Z}\rightarrow \mathcal{A}$ attains CDP if  $\mathcal{D}(\boldsymbol{Z})$  satisfies
\begin{equation}
\label{e5}
\begin{aligned}
{P}[\mathcal{D}(\boldsymbol{Z})=a \mid {S}={s}, L=l]={P}[\mathcal{D}(\boldsymbol{Z})=a\mid {S}={s}^{\prime},L=l],
\nonumber
\end{aligned}
\end{equation}
for any  $ {s}, {s}^{\prime} \in \mathcal{S}$, $l \in \mathcal{L}$ and $a \in \{-1,1\}$ .
\end{definition}
{This definition requires that IDRs be  independent of  $S$ given a ``legitimate'' feature $L$, allowing decision rules to vary across groups $l$ in decision-making.
For instance, in a credit lending scenario, each applicant is assigned a credit grade $L$. The CDP-IDR implies that applicants with the same credit grade should have an equal probability of loan approval, irrespective of $S$. The choice of  $L$  is determined by policymakers and reflects permissible difference for decision-making.  $L$ can be constructed by stratifying the population based on  some ``legitimate'' factors (such as  income, savings, or credit history), thereby ensuring that the fairness constraint is enforced only within ethically and legally acceptable groups.}

\section{Methodology}

\subsection{Conditional Demographic Parity IDR}
We derive the optimal IDR under the CDP constraint, where the DP-IDR corresponds to a special case when  the entire population is treated as a single group, i.e.,  $L$ is a constant.
The optimal CDP-IDR is denoted by  $\mathcal{D}_{cdp}^{*}(\boldsymbol{Z})$, which solves the following optimization problem:
\begin{equation}
\begin{aligned}
\label{e.8}
 \max_{\mathcal{D}} \quad &  \mathcal{V}(\mathcal{D}),\\
 \text { subject to }&
{P}[\mathcal{D}(\boldsymbol{Z})=a \mid {S}={s}, L=l]={P}[\mathcal{D}(\boldsymbol{Z})=a\mid {S}={s}^{\prime},L=l],
\end{aligned}
\end{equation}
for any  $ {s}, {s}^{\prime} \in \mathcal{S}$, $l \in \mathcal{L}$  and $a \in \{-1,1\}$.

In view of
\begin{equation}
\begin{aligned}
\label{constraint_tansforma}
{P}[\mathcal{D}(\boldsymbol{Z})=1 \mid {S}=s, L=l]&=\frac{{P}[f(\boldsymbol{Z})>0, {S}=s,L=l]}{P[{S}=s, L=l]}\\
&=E\left\{\frac{I(f(\boldsymbol{Z})>0)I \left(
{S}=s,L=l  \right)}{P[{S}=s, L=l]}\right\},
\end{aligned}
\end{equation}
the CDP constraint in Equation (\ref{e.8}) is equivalent to
\begin{equation}
\begin{aligned}
E\left[I(f(\boldsymbol{Z})>0)I(L=l) \left\{\frac{ I\left(
{S}=1  \right)}{\pi(1 | l)}-\frac{ I\left(
{S}=0 \right)}{\pi(0 | l)} \right\}\right]=0,  \text {  for any } l \in \mathcal{L},
\nonumber
\end{aligned}
\end{equation}
where $\pi(s | l):= P({S}=s | L=l)$.
Therefore,
Equation (\ref{e.8}) is rewritten as
\begin{equation}
\begin{aligned}
\label{e.4}
 \max_{\mathcal{D}} \quad &   E \left\{I(f(\boldsymbol{Z})>0)\delta_R(\boldsymbol{Z})\right\},\\
 \text {subject to }&
E\left[I(f(\boldsymbol{Z})>0)I(L=l) \left\{\frac{ I\left(
{S}=1  \right)}{\pi(1 | l)}-\frac{ I\left(
{S}=0 \right)}{\pi(0 | l)} \right\}\right]=0,  \text {  for any } l \in \mathcal{L}.
\end{aligned}
\end{equation}

\begin{proposition}
Equation (\ref{e.8}) has the solution $\mathcal{D}_{cdp}^{*}(\boldsymbol{Z}):=\sum_{l\in\mathcal{L}} \mathcal{D}_l^{*}(\boldsymbol{Z})I(L=l),$ where $$\mathcal{D}_l^{*}=2I(f_l^{*}(\boldsymbol{Z})>0)-1$$  and $f_l^{*}(\boldsymbol{Z})$  solves
\begin{equation}
\begin{aligned}
\label{e.9}
 \max_f  & E\left[I(f(\boldsymbol{Z})>0)\delta_R(\boldsymbol{Z}) \right]\\
 \text{ subject to }
& E\left[I(f(\boldsymbol{Z})>0)I(L=l) \left\{\frac{ I\left(
{S}=1  \right)}{\pi(1 | l)}-\frac{ I\left(
{S}=0\right)}{\pi(0 | l)} \right\} \right]=0,
\end{aligned}
\end{equation}
where $\pi(s | l)=P({S}=s | L=l)$.
\end{proposition}
Actually, the IDR $\mathcal{D}_l^{*}$ satisfies demographic parity in the $l$-th group. To solve the problem (\ref{e.9}), we borrow
the idea of \cite{risk2018}. After introducing the Lagrange multiplier, the problem (\ref{e.9})  is transformed into
\begin{equation}
\begin{aligned}
\label{e.6}
   &\max_f E\left[I(f(\boldsymbol{Z})>0)\{\delta_R(\boldsymbol{Z}) -\omega_{l} I(L=l)\psi_l(S) \}\right]\\
 &\text{subject to }
 E\left[I(f(\boldsymbol{Z})>0)I(L=l)\psi_l(S) \right]=0,
\end{aligned}
\end{equation}
where $\psi_l(S):=  \frac{I(S = 1)}{\pi(1 | l)} - \frac{I(S = 0)}{\pi(0 | l)}$ and $\omega_{l}$ is a Lagrange multiplier. Therefore,
the solution $f^{*}_{l}$  satisfies that $$I[f^{*}_{l}(\boldsymbol{Z})>0]=I[\delta_R(\boldsymbol{Z})-\omega_{l}^{*} I(L=l)\psi_l(S)>0],$$ where $\omega_{l}^{*}$ is the zero solution of equation $ E\left[I\{\delta_R(\boldsymbol{Z})-\omega_{l}^{*} \psi_l(S)>0\}I(L=l)\psi_l(S) \right]=0$.

Note that
\begin{equation}
\begin{aligned}
&E\left[I\{\delta_R(\boldsymbol{Z})-\omega_{l} \psi_l(S)>0\}I(L=l)\psi_l(S) \right]\\
=&E\left[I\left\{ \psi_l(S)>0,\frac{\delta_R(\boldsymbol{Z})}{\psi_l(S)}>
\omega_{l} \right \}I(L=l)\psi_l(S) \right]\\
+&E\left[I\left\{\psi_l(S)<0,\frac{\delta_R(\boldsymbol{Z})}{\psi_l(S)}<
\omega_{l}\right \}I(L=l)\psi_l(S) \right]
\nonumber
\end{aligned}
\end{equation}
is non-increasing in  $\omega_{l}$, guaranteeing a unique zero solution $\omega_{l}^*$. Therefore, we have $$\mathcal{D}_l^{*}(\boldsymbol{Z})=2I\big\{\delta_R(\boldsymbol{Z})-
\omega_{l}^*I(L=l)\psi_l(S)>0\big\}-1,$$ and
the CDP-IDR is $$\mathcal{D}^{*}_{cdp}(\boldsymbol{Z})=2I\big\{\delta_R(\boldsymbol{Z})-
\sum_{l\in \mathcal{L}}I(L=l)\omega_{l}^*\psi_l(S)>0\big\}-1.$$ This expression provides a closed-form characterization of the CDP-IDR.

\subsection{Trade-off between Conditional Demographic Parity and Value Optimality}
In many practical applications, policymakers must tailor the level of fairness to align with the objectives of  specific programs. Therefore, we propose the $\epsilon$-CDP-IDR, which allows a controlled deviation from CDP  to balance between value optimality and fairness constraint.

Specifically, we define the $\epsilon$-CDP-IDR, denoted by $\mathcal{D}_{cdp}^{\epsilon}(\boldsymbol{Z})$, as the solution to the following constrained optimization problem:
\begin{equation}
\begin{aligned}
\label{e_section_3.4}
 \max_{\mathcal{D}} \quad &   \mathcal{V}(\mathcal{D}),\\
 \text { subject to }&
\big|{P}[\mathcal{D}(\boldsymbol{Z})=a \mid {S}={s}, L=l]-{P}[\mathcal{D}(\boldsymbol{Z})=a\mid {S}={s}^{\prime},L=l]\big|\leq \epsilon
\end{aligned}
\end{equation}
for any  $ {s}, {s}^{\prime} \in \mathcal{S}$ and $l \in \mathcal{L}$.
The constraint enforces $\epsilon$-level DP  within the $l$-th group.
Here,
$\epsilon$ is a constraint specified by decision makers, which represents the acceptable tolerance for unfairness. When $\epsilon=0$, $\epsilon$-CDP-IDRs correspond to CDP-IDRs.

The following proposition provides an explicit construction of  $\epsilon$-CDP-IDR.
\begin{proposition}
The solution to problem \eqref{e_section_3.4} admits a partitioned structure:
\[
\mathcal{D}_{cdp}^{\epsilon}(\boldsymbol{Z}) = \sum_{l \in \mathcal{L}} \mathcal{D}_l^{\epsilon}(\boldsymbol{Z}) \cdot I(L = l),
\]
where the subgroup-specific IDR $\mathcal{D}_l^{\epsilon}$ is defined as
$
\mathcal{D}_l^{\epsilon}(\boldsymbol{Z}) = 2I\left(f_l^{\epsilon}(\boldsymbol{Z}) > 0\right) - 1
$
and $f_l^{\epsilon}$ solves the constrained optimization problem:
\begin{equation}
\begin{aligned}
\label{e_section_3.4_proposition}
 \max_f \quad & E\left[I(f(\boldsymbol{Z}) > 0)  \delta_R(\boldsymbol{Z})\right]\\
 \text{subject to} \quad &
 \left|E[I(f(\boldsymbol{Z}) > 0)  I(L = l)  \psi_l(S)] \right| \leq \epsilon,
\end{aligned}
\end{equation}
with $\psi_l(S)= \frac{I(S = 1)}{\pi(1 | l)} - \frac{I(S = 0)}{\pi(0 | l)}$, and $\pi(s | l)= P(S = s \mid L = l)$.
\end{proposition}
This optimization problem can also be expressed via the following pair of  inequality constraints:
\begin{equation}
\begin{aligned}
\label{discussions_24}
 \max_f \quad & E\left[I\{f(\boldsymbol{Z})>0\}\delta_R(\boldsymbol{Z})
  \right],\\
 \text { subject to }
&E [I(f(\boldsymbol{Z}) > 0)  I(L = l)  \psi_l(S)] -\epsilon \leq 0,\\
-&E [I(f(\boldsymbol{Z}) > 0)  I(L = l)  \psi_l(S)]-\epsilon \leq 0.
\nonumber
\end{aligned}
\end{equation}
We first introduce Lagrange multipliers and obtain the corresponding Lagrangian function:
\begin{equation}
\begin{aligned}
\label{discussions_25}
E\left[I\{f(\boldsymbol{Z})>0\}\{\delta_R(\boldsymbol{Z})-(\lambda-\mu)I(L = l)  \psi_l(S)
 \} \right]-\epsilon(\lambda+\mu), \quad
\nonumber
\end{aligned}
\end{equation}
where $\lambda \geq0, \mu\geq0$. Let $\omega:=\lambda-\mu$ and $\eta:=\lambda+\mu$. We hence focus on solving the equation:
\begin{equation}
\begin{aligned}
\label{e.11}
f_{l,\omega}^{*} := \arg \max_f  E\left[I\{f(\boldsymbol{Z})>0\}\{\delta_R(\boldsymbol{Z})- \omega I(L = l)  \psi_l(S)
 \} \right],
 \nonumber
\end{aligned}
\end{equation}
for each $\omega$,
where $\eta$ is omitted because it is irrelevant to $f$. Consequently, we obtain $I\{f_{l,\omega}^{*} (\boldsymbol{Z})>0\}=I\{\delta_R(\boldsymbol{Z})-\omega I(L = l)  \psi_l(S) >0 \}.$

Next, we show that there exists a $\omega^{*}_{l}$ such  that $f_{l, \omega^{*}_{l}}^{*}$ is the solution of original optimization problem (\ref{e_section_3.4_proposition}).
Given $f_{l, \omega}^{*}$, we define the   constraint function and value function:
\begin{equation}
\begin{aligned}
G_l(\omega):=  E\left[I(f_{l, \omega}^{*} (\boldsymbol{Z})>0) I(L = l)  \psi_l(S)\right], \quad
V_{l}(\omega):=  E\left[I(f_{l, \omega}^{*} (\boldsymbol{Z})>0)\delta_R(\boldsymbol{Z})\right].
\nonumber
\end{aligned}
\end{equation}
\begin{lemma}
Assume that $G_l$ is continuous for $\omega \in \mathbb{R}$ and Conditions (C1) and (C4) hold. Then the following properties hold:
\begin{itemize}
    \item $G_l(\omega)$ is non-increasing for all $\omega \in \mathbb{R}$;
        \item $V_{l}(\omega)$ is non-decreasing for $\omega \leq 0$ and non-increasing for $\omega > 0$.
\end{itemize}
Moreover, for a sufficiently large constant $K > 0$:
\begin{itemize}
    \item If $G_l(0) > \epsilon$, then there exists $\omega^{*}_{l} \in (0, K)$ such that $G_l(\omega^{*}_{l}) = \epsilon$;
    \item If $G_l(0) < -\epsilon$, then there exists $\omega^{*}_{l} \in (-K, 0)$ such that $G_l(\omega^{*}_{l}) = -\epsilon$.
\end{itemize}
\end{lemma}
Lemma 1 implies that the expected value function $V_{l}(\omega)$ achieves its maximum at $\omega = 0$. If the constraint is satisfied  when $\omega = 0$, i.e., $|G_l(0)| \leq \epsilon$, then $f_{l,0}^{*}$ is the optimal solution of (\ref{e_section_3.4_proposition}).
Otherwise, consider the case where $G_l(0) > \epsilon$.
 Let $\omega^{*}_{l} \in (0, K)$ satisfy $G_{l}(\omega^{*}_{l}) = \epsilon$. For any $\omega$ satisfying $|G_l(\omega)| \leq \epsilon$, based on the monotonicity of $G_l(\omega)$ and $V_{l}(\omega)$, it follows  that $\omega \leq \omega^{*}_{l}$ and $V_{l}(\omega) \leq V_{l}(\omega^{*}_{l})$.
 Our main result in the following theorem establishes that $f_{l, \omega^{*}_{l}}^{*}$ is indeed the optimal solution to the constrained optimization problem (\ref{e_section_3.4_proposition}).
\begin{theorem} Suppose that,  for each $l\in \mathcal{L}$, $G_l(\omega)$ is continuous in $\omega$. Define the optimal Lagrange multiplier $\omega^*_l$ as:
\[
\omega^{*}_l =
\begin{cases}
0, & \text{if } |G_l(0)| \leq \epsilon, \\
\text{such that } G_l(\omega^{*}_l) = \epsilon, & \text{if } G_l(0) > \epsilon, \\
\text{such that } G_l(\omega^{*}_l) = -\epsilon, & \text{if } G_l(0) < -\epsilon.
\end{cases}
\]
Then,  for any $f$ satisfying the constraint  $|E\left[I\{f(\boldsymbol{Z})>0\}I(L=l)\psi_{l}(S)\right]| \leq \epsilon$, it holds that $
E\left[\delta_R(\boldsymbol{Z}) I\{f(\boldsymbol{Z})>0\}\right]\leq  E\left[\delta_R(\boldsymbol{Z}) I\{f_{l, \omega^{*}_l}^{*}(\boldsymbol{Z})>0\}\right]$.
Consequently,  the  $\epsilon$-CDP-IDR can be written as $$\mathcal{D}^{\epsilon}_{cdp}(\boldsymbol{Z})=2I\{\delta_R(\boldsymbol{Z})-\sum_{l\in\mathcal{L}} I(L=l)\omega^{*}_l  \psi_l(S)
 >0\}-1.$$
\end{theorem}
This result establishes that the  $\epsilon$-CDP-IDR can be derived by solving a one-dimensional root-finding problem over the scalar $\omega$, which can be efficiently computed using numerical methods such as the bisection algorithm \citep{solanki2014role} over a bounded interval $[-K, K]$.

Moreover, the explicit form of $\mathcal{D}^{\epsilon}_{cdp}$ implies that enforcing a conditional demographic parity constraint is equivalent to perturbing the unconstrained optimal IDR by introducing a fairness-aware perturbation term $\sum_{l\in\mathcal{L}} I(L=l)\omega^{*}_l  \psi_l(S)$. The scalar $\omega^{*}_l$ quantifies the magnitude of perturbation  while $G_l(0)$ determines its direction.

The fairness-aware perturbation term $\sum_{l\in\mathcal{L}} I(L=l)\omega^{*}_l  \psi_{l}(S)$ is group-specific, with each subgroup $L=l$ assigned its specific perturbation level $\omega^{*}_l$.  Specially, the $\epsilon$-DP-IDR can be easily obtained by setting $L$ to be a constant assigned to every individual. Therefore, the $\epsilon$-DP-IDR applies a uniform perturbation term across all subgroups. This distinction allows $\epsilon$-CDP-IDR to more flexibly balance reward maximization and fairness constraints—especially in settings where the degree of unfairness varies substantially across groups.

\section{Estimation Procedure}
Consider the observed dataset $\left\{ (\boldsymbol{X}_i,  S_i, L_i, A_i, R_i) \right\}_{i=1}^n$, which consists of  two subsamples: $A_i = 1$ for $i = 1, 2, \ldots, n_1$ and  $A_i = -1$ for $i = n_1+1, n_1+2, \ldots, n_1+n_0$, where $n = n_1 + n_0$.  Let $N_l$ denote the number of observations with $L_i=l$ and $N_1\leq N_2, \ldots, \leq N_{|\mathcal{L}|}$, where $|\mathcal{L}|$ denotes the cardinality of set $\mathcal{L}$. Then, we have $\sum_{l\in \mathcal{L}} N_l=n$.

We propose  the following estimation procedure for $\mathcal{D}^{*}_{cdp}$ and ${\mathcal{D}}_{cdp}^{\epsilon}$.
\begin{itemize}
\item  Step 1: We estimate $E(R| \boldsymbol{Z}, A=1)$ and $E(R| \boldsymbol{Z}, A=-1)$ by minimizing the mean squared error loss for the respective groups $A=1$ and $A=-1$:
\begin{equation}
\begin{aligned}
&\widehat{m}_1:= \arg \min _{f \in \mathcal{F}_1}  \sum_{i=1}^{n_1} \{(R_i-f(\boldsymbol{Z}_i))^2  \}/n_1, \\
& \widehat{m}_{0}:= \arg \min _{f \in \mathcal{F}_0 } \sum_{i=n_1+1}^{n} \{(R_i-f(\boldsymbol{Z}_i))^2 \}/n_0,
\nonumber
\end{aligned}
\end{equation} where  $ \mathcal{F}_{j}$, $j=0, 1$,  denotes the class of feedforward neural networks with rectified linear unit (ReLU) activation function.
Each function $f \in \mathcal{F}_{j}$,  $j=0, 1$, subject to $\left\|f\right\|_{\infty} \leq \mathcal{B}$, can be represented as:
$$
f(x)=\mathcal{L}_{\mathcal{D}_{j}} \circ \sigma \circ \mathcal{L}_{\mathcal{D}_{j}-1} \circ \sigma \circ \cdots \circ \sigma \circ \mathcal{L}_1 \circ \sigma \circ \mathcal{L}_0(x), \quad x \in \mathbb{R}^{d},
$$
where $\sigma(x)=\max (0, x)$ is the ReLU activation function and $\mathcal{L}_i(x)=W_i x+b_i, i=0,1, \ldots, \mathcal{D}_{j}$, where $W_i \in \mathbb{R}^{p_{i+1} \times p_i}$ is a weight matrix, $p_i$ is the width of the $i$th layer, and $b_i \in \mathbb{R}^{p_{i+1}}$.
The network has $\mathcal{D}_{j}$ hidden layers; width $\mathcal{W}_{j} = \max\{p_1, \ldots, p_{\mathcal{D}_{j}}\}$;  the total number of parameters in the network $\mathcal{S}_{j} = \sum_{i=0}^{\mathcal{D}_{j}} p_{i+1}(p_i + 1)$; and number of neurons $\mathcal{U}_{j} = \sum_{i=1}^{\mathcal{D}_{j}} p_i$. For $j=0$ or $1$, The class $\mathcal{F}_{j}$ is thus characterized by its depth $\mathcal{D}_{j}$, width $\mathcal{W}_{j}$, and size $\mathcal{S}_{j}$.
We then obtain the  estimator for CATE  $\widehat{\delta}_R(\boldsymbol{Z})=\widehat{m}_1(\boldsymbol{Z})-\widehat{m}_0(\boldsymbol{Z})$.

\item Step 2:
We estimate  $P(S=1|L=l)$ and  $P(S=0|L=l)$ with $ \widehat{\pi}_1(l)=\sum_{i=1}^{n}I(S_i=1, L_i=l)/N_l$, $  \widehat{\pi}_0(l)=\sum_{i=1}^{n}I(S_i=0, L_i=l)/N_l$, respectively.
Subsequently, we obtain $\widehat{\psi}_l(S)= I(S=1)/\widehat{\pi}_1(l)-I(S=0)/\widehat{\pi}_0(l)$.

\item Step 3:
To address the potential discontinuity of the constraint function $G_l$,
we consider a smooth approximation given by
$$\widehat{G}_l(\omega) = n^{-1}\sum_{i=1}^{n} \left[\widehat{\psi}_{l}(S_{i})I(L_i=l)\Phi\left(\frac{\widehat{\delta}_R(\boldsymbol{Z}_{i})-\omega \widehat{\psi}_l(S_{i})}{h}\right)\right],$$
where $\Phi$ is the cumulative distribution function of   the standard normal distribution and $h>0$ is a small bandwidth parameter.
We then estimate $\widehat{\omega}_{l}^{\epsilon}$ by solving the equation
$$
\widehat{G}_l(\omega) =
\begin{cases}
\epsilon, & \text{if } \displaystyle \frac{1}{n} \sum_{i=1}^{n} \left[\widehat{\psi}_l(S_{i})I(L_i=l)I\left(\widehat{\delta}_R(\boldsymbol{Z}_{i}) > 0\right)\right] > \epsilon, \\
-\epsilon, & \text{if } \displaystyle \frac{1}{n} \sum_{i=1}^{n} \left[\widehat{\psi}_l(S_{i})I(L_i=l)I\left(\widehat{\delta}_R(\boldsymbol{Z}_{i}) > 0\right)\right] <-\epsilon, \\
0, & \text{otherwise}.
\end{cases}
$$
We use the bisection method \citep{solanki2014role} in an interval $[-K, K]$ to search the root.
\item Step 4:
Finally, we obtain the estimator of $\epsilon$-CDP-IDR: $$\widehat{\mathcal{D}}_{cdp}^{\epsilon}(\boldsymbol{Z})=2I\big\{ \widehat{\delta}_R(\boldsymbol{Z})-\sum_{l\in\mathcal{L}} I(L=l)\widehat{\omega}^{\epsilon}_l  \widehat{\psi}_{l}(S)>0\big\}-1.$$  When we set $\epsilon=0$, we obtain the estimator of CDP-IDR $$\widehat{\mathcal{D}}_{cdp}(\boldsymbol{Z})=2I\big\{ \widehat{\delta}_R(\boldsymbol{Z})-\sum_{l\in\mathcal{L}} I(L=l)\widehat{\omega}^{0}_l  \widehat{\psi}_{l}(S)>0\big\}-1.$$  We write $\widehat{\omega}^{\epsilon}_{l}$ as $\widehat{\omega}_l$ to simplify notation.
\end{itemize}

\section{Theoretical Results}
Let  $\mathcal{D}^{*}$  be the unconstrained optimal IDR.
We aim to obtain an upper bound for  $\mathcal{V}(\mathcal{D}^{*})-\mathcal{V}(\widehat{\mathcal{D}}_{cdp}^{\epsilon})$.
Let $\widehat{f}_{\widehat{\boldsymbol{\omega}}}(\boldsymbol{Z})=\widehat{\delta}_R(\boldsymbol{Z})-\sum_{l\in\mathcal{L}} I(L=l)\widehat{\omega}_l  \widehat{\psi}_{l}(S)$, where  $\widehat{\boldsymbol{\omega}}=(\widehat{\omega}_{1}, \ldots,\widehat{\omega}_{|\mathcal{L}|})$.
Define  $\mathcal{G}_l(\widehat{f}_{\widehat{\boldsymbol{\omega}}}):=E_{\boldsymbol{Z}}\left[I\left\{\widehat{f}_{\widehat{\boldsymbol{\omega}}}(\boldsymbol{Z})>0 \right\} I(L=l)\psi_{l}(S)\right].$ We further derive upper bounds  for the CDP constraints $|\mathcal{G}_l(\widehat{f}_{\widehat{\boldsymbol{\omega}}})|-\epsilon$, for $l \in \mathcal{L}$.
To establish the theoretical properties of the proposed estimators, we need the following regularity conditions.
\begin{enumerate}
  \renewcommand{\labelenumi}{(C\arabic{enumi})}
  \renewcommand{\theenumi}{C\arabic{enumi}}
 \setlength{\itemindent}{11pt} % 设置列表项的缩进为0
  \setlength{\labelsep}{5pt} % 设置标签与文本之间的间距为0
  \setlength{\labelwidth}{0pt} % 设置标签的宽度为0
  \setlength{\leftmargin}{0pt} %
\item   For any $s \in \mathcal{S}$ and $l \in \mathcal{L}$, we have  $P(S = s| L=l) > c_0>0$.
\item The propensity score $e(\boldsymbol{Z}) = P(A = a | \boldsymbol{Z}) > c_1 > 0$, for $a \in \mathcal{A}.$
\item The reward $R$ is sub-exponentially distributed, that is, there exists a constant $\sigma_R>0$ such that ${E} \exp \left(\sigma_R|R|\right)<\infty$.
\item  The true functions $f_0(\boldsymbol{z}):=E(R|\boldsymbol{Z}=\boldsymbol{z}, A=-1)$ and $f_1(\boldsymbol{z}):=E(R|\boldsymbol{Z}=\boldsymbol{z}, A=1)$  belong to a Hölder class $\mathcal{H}^\beta\left([0,1]^d, B_0\right)$,  defined as {\small
$$
\begin{aligned}
 \mathcal{H}^\beta\left([0,1]^d, B_0\right)=\left\{f:[0,1]^d \rightarrow \mathbb{R}, \max _{\|\boldsymbol{ \alpha}\|_1 \leq s}\left\|\partial^{\boldsymbol{ \alpha}} f\right\|_{\infty} \leq B_0, \max _{\|\boldsymbol{ \alpha}\|_1=s} \sup _{x \neq y} \frac{\left|\partial^{\boldsymbol{ \alpha}} f(x)-\partial^{\boldsymbol{ \alpha}} f(y)\right|}{\|x-y\|_2^r} \leq B_0\right\},
\end{aligned}
$$}
where $d=p+2$, $\partial^{\boldsymbol{ \alpha}} = \partial^{\alpha_1} \dots \partial^{\alpha_d}$ with $\boldsymbol{ \alpha} = (\alpha_1, \dots, \alpha_d)^{\top} \in \mathbb{N}_0^d$ and $\|\boldsymbol{ \alpha}\|_1 = \sum_{i=1}^d \alpha_i$. Here, $\beta = s + r > 0$, $r \in (0,1]$, and $s = \lfloor \beta \rfloor \in \mathbb{N}_0$, where $\lfloor \beta \rfloor$ is the largest integer strictly smaller than $\beta$ and $\mathbb{N}_0$ denotes the set of nonnegative integers.
\item The conditional density function of ${\delta}_R(\boldsymbol{Z})$ given $S = s, L=l$ is differentiable.
\item There exist constants $m_0>0$ and   $\eta \geq0$ such that
$
P\left(|\delta_R(\boldsymbol Z)| \le  \gamma\right) \le m_0\,  \gamma^\eta
$ for all positive $\gamma$.
\end{enumerate}
Conditions (C1) and (C2) ensure that there are sufficient samples for different sensitive attributes and treatments.
Condition (C3) requires that the tail of the distribution $R$ decay rapidly.  Under Condition (C4), the target function is a smooth function, which is standard condition in DNNs literature  \citep{jiao2023deep, yan2025deep}. From a theoretical perspective, when discrete covariates take values in a finite set, the true conditional mean function can be estimated given those covariates. In practice, deep neural networks have demonstrated strong empirical performance in handling such discrete inputs, as shown in \citet{farrell2021deep}.  {Condition (C5) is a technical assumption for establishing the convergence rate of the estimated smooth function $\widehat{G}_l$.}
Condition (C6) is a standard assumption, which has been used in \citet{qian2011performance} and \citet{shi2020breaking}.

\begin{theorem}
Suppose that Conditions (C1)-(C5) and $\mathcal{B} \geq$ $\max \left\{1, B_0\right\}$  hold.  Consider the function classes of feedforward neural networks $\mathcal{F}_{j}$,  $j=0, 1$, where the depth $\mathcal{D}_{j}$, width $\mathcal{W}_{j}$ and size $\mathcal{S}_{j}$ are given by
$
\mathcal{W}_j  =O\left(n_{j}^{d / 4(d+2 \beta)} \log _2(n_{j})\right), \quad \mathcal{D}_j=O\left(n_{j}^{d / 4(d+2 \beta)} \log _2(n_{j})\right),
\mathcal{S}_j  =O\left(n_{j}^{3 d / 4(d+2 \beta)}(\log n_{j})^4\right)
$.
For  sufficiently small $\delta>0$ and sufficiently large $N_1$, we have
\begin{equation}
\begin{aligned}
\label{e15}
\mathcal{V}(\mathcal{D}^{*})-\mathcal{V}(\widehat{\mathcal{D}}_{cdp}^{\epsilon}) &\leq
O(h^2)+O_p(N_{1}^{-1/2}) +  \frac{\bar{c}}{c_0}\max_{l\in\mathcal{L}}|{\omega}_l^{*}| \\
&+c_4 c_3(\delta,\mathcal{B},d,\beta)\left \{n_1^{- \beta /(d+2 \beta)}(\log n_1)^{11/2}+n_0^{- \beta /(d+2 \beta)}(\log n_0)^{11/2} \right\}\\
&+ c_5\delta^{-1/2} N_{1}^{-1/2},
\end{aligned}
\end{equation}
with probability greater than $1-\delta$,  where  $ \bar{c}=\frac{E(|R|)}{c_1}$, $c_3(\delta,\mathcal{B},d,\beta)=\sqrt {c_2 }(\sqrt{14/\delta}+1)\mathcal{B}^{5/2}(\lfloor\beta\rfloor+1)^2 d^{\lfloor\beta\rfloor+\max\{\beta, 1\}/2}$, and $c_2$,  $c_4$ and $c_5$ are constants.
\end{theorem}
In Theorem 2, as $N_1,n_1,n_0 \to \infty$ and $h \to 0$,  the dominant source of value loss is the term $\frac{\bar{c}}{c_0}\max_{l\in\mathcal{L}}|{\omega}_l^{*}|$. In particular, $\max_{l \in \mathcal{L}} |\omega_l^{*}|$ reflects the maximum perturbation required to enforce  demographic parity across groups.  Notably, for a given tolerance level $\epsilon$, if the optimal IDR satisfies the CDP constraint, i.e., $|G_l(0)| \leq \epsilon$ for all $l \in \mathcal{L}$, then $\omega_l^{*} = 0$, implying that no value loss is incurred under the CDP constraint.

\begin{theorem}
Suppose that Conditions (C1)-(C5)  and the conditions in Theorem 2 hold. For any $l\in\mathcal{L}$ and a sufficiently small $\delta > 0$  and sufficiently large $N_1$, we have
\begin{equation}
\begin{aligned}\label{J.7}
|\mathcal{G}_{l}(\widehat{f}_{\widehat{\boldsymbol{\omega}}})|
&\leq  \epsilon+ O(h^{2})
+O_p(\frac{1}{h}N_1^{-1/2})
+\frac{c_{6}}{h}\delta^{-1/2}N_1^{-1/2}
\\&+ \left\{ \frac{c_7c_3(\delta,\mathcal{B},d,\beta)}{h} +\textcolor{red}{2 c_8\sqrt{\frac{14}{\delta}}} \right\}
\left \{n_1^{- \beta /(d+2 \beta)}(\log n_1)^{11/2}+n_0^{- \beta /(d+2 \beta)}(\log n_0)^{11/2} \right\}^{\frac{\eta}{(1+\eta)}} ,
\end{aligned}
\end{equation}
with probability greater than $1-\delta$, where $c_6$,  $c_7$and $c_{8}$ are constants, and $c_3(\delta,\mathcal{B},d,\beta)$ is defined in  Theorem 2.
\end{theorem}
Theorem 3 shows that the CDP constraint can be satisfied asymptotically as $h \to 0$ and $n \to \infty$.
In particular, assume $n_1 = n_0 = n/2$, $N_1=n/|\mathcal{L}|$ and $\eta =1$. Then we obtain
$|\mathcal{G}_{l}(\widehat{f}_{\widehat{\boldsymbol{\omega}}})| \leq  \epsilon+ O(h^{2})+O_p({h}^{-1}n^{- \beta /\{2(d+2 \beta)\}})$.  Choosing the  bandwidth $h = n^{-\beta /\{6(d+2\beta)\}}$, we also achieve the rate  $|\mathcal{G}_{l}(\widehat{f}_{\widehat{\boldsymbol{\omega}}})| \leq \epsilon+ O_p(n^{- \beta /\{3(d+2 \beta)\}})$.

\section{Simulation Studies}

\subsection{Simulation Designs}
We present four simulation cases to evaluate the finite-sample performance of the proposed  DP-IDR, $\epsilon$-DP-IDR, CDP-IDR, and $\epsilon$-CDP-IDR.

For Cases 1 and 2,
 the response variable $R$  was modeled as:
$$R=T_0(\boldsymbol{X}, S, A)+ \xi, \quad \xi \sim \mathcal{N}(0,1),$$  where  $\xi$  was independent of $ (\boldsymbol{X}, S)$.

For Cases 3 and 4,  the response variable $R$  was modeled as: $$R=T_1(\boldsymbol{X},S, L, A)+ \xi,  \quad  \xi \sim \mathcal{N}(0,1),$$  where  $\xi$ was independent of $ (\boldsymbol{X}, S, L)$.

For all four cases, the components of
$\boldsymbol{X}$ were independently drawn from a normal distribution  $\mathcal{N}(0,1)$ and  truncated by $[-10, 10]$ with dimensions $p=3, 3, 30 $, and $ 30$,  respectively.  The sensitive attribute $S$ was generated from  a Bernoulli distribution with  $P(S=1 |\boldsymbol{X} )=\frac{ X_1^{2} }{ 2X_1^{2}+  X_2^{2}}$, where $X_1$ and $X_2$ denoted the first and second components of $\boldsymbol{X}$, respectively. For Cases 3 and 4,  the binary variable $ L $ was generated from a Bernoulli distribution with
$P(L = 1 \mid \boldsymbol{X}, S) = \frac{1}{1 + \exp\{ (1-2X_3+X_4-2X_5^{2} - S) \} }.$
The treatment $A$ was drawn from $\{-1,1\}$ with equal probability.
We considered the following different $T_0$ and $T_1$:

\noindent\textbf{Case\ 1:}
$T_0(\boldsymbol{X}, S, A)=\left\{ \left(|X_1-X_2|+0.5\right) \operatorname{sign}(X_1-X_2)+S\right\}A.$

\noindent\textbf{Case\ 2:}
$T_0(\boldsymbol{X}, S, A)=\left\{S(X_1-X_{2})^2+0.5) \operatorname{sign}(X_1-X_2^2)+S\right\}A,$

\noindent\textbf{Case\ 4:}
$T_1(\boldsymbol{X}, S, L,A)=\left\{X_1-X_2^{2}+ \sin(X_3X_4) + \ln{(|X_5|+0.1)}-2S+L\right\}A.$

\noindent\textbf{Case\ 5:}
$T_1(\boldsymbol{X}, S, L, A)=\left(X_1X_2+ \exp{X_3} +|X_4|+X_5+2S+L\right)A.$

The dataset was partitioned into three sets: the training and validation sets together that contained $n$ samples ($4n/5$ for training and $n/5$ for validation), and the testing set contained $1000$ samples.    These cases were repeated 200 times.
We assessed the performance of the  estimated IDR using two metrics: the unfairness level (UF) and the  policy value (PV). Specifically,
for any an estimated decision function $\widehat{f}$ and the corresponding IDR defined as $ \widehat{\mathcal{D}}(\boldsymbol{Z})=2I(\widehat{f}(\boldsymbol{Z})>0)-1$,
the unfairness level was given by
$$\left|\frac{\sum_{j=1}^{N}I(\widehat{\mathcal{D}}(\boldsymbol{Z}_{j})=1)I(S_{j}=1)}{N_1}- \frac{\sum_{j=1}^{N}I(\widehat{\mathcal{D}}(\boldsymbol{Z}_{j})=0)I(S_{j}=0)}{N_0}\right|,$$ where $\boldsymbol{Z}_{j}, j=1, 2, \ldots, N,$ were samples from the testing set.  $N_{1}$ and $N_0$ denoted the number of samples with $S=0$ and $S=1$, respectively. This  unfairness level calculated the absolute difference in the proportion of positive decisions between the two groups.
The empirical policy value   was defined as $\frac{1}{N}\sum_{j=1}^{N} \delta_R(\boldsymbol{Z}_j)I(\widehat{f}(\boldsymbol{Z}_j)>0)$, where $\delta_R$ was given by the real  model.

The performance of CDP-IDR was assessed using the conditional unfairness level (CUF) $\frac{1}{|\mathcal{L}|} \sum_{l\in \mathcal{L}} \text{UF}(l), $
where  $|\mathcal{L}|$ denoted the cardinality of the set $ \mathcal{L}$ and  $\text{UF}(l)$ was the unfairness level for the $l$-th group.

We compared our proposed method with two  approaches.

{Fair CATE} \citep{kim2023fair}: This method estimated the conditional average treatment effect, denoted by $\widehat{\delta}_R^{1}(\boldsymbol{X}, S)$, under the demographic parity constraint. The corresponding estimator  of IDR was defined as $2I\{\widehat{\delta}_R^{1}(\boldsymbol{X}, S) > 0\} - 1$.

{AF-IDR} \citep{GITR}:  { The method learned a representation  that satisfied the demographic parity constraint.  Then, the learned representation  was used to estimate the conditional average treatment effect, denoted by $\widehat{\delta}_R^{2}(\boldsymbol{X})$, via  a deep neural network (see Step 1 in Section 4.1 for details). The estimator of IDR was given  by $2I\{\widehat{\delta}_R^{2}(\boldsymbol{X}) > 0\} - 1$.}

\subsection{Simulation Results}
\begin{table*}[h]
\caption{Sample means and sample standard deviations (in parentheses) of the unfairness levels (UF) and policy values (PV) for the proposed DP-IDR, Fair CATE and AF-IDR methods, based on 200 replications under Case 1.}
  \centering
\begin{tabular}{clcccc}
   \toprule
 & Method  & n=500 & n=1000 & n=2000 \\
       \hline
\multirow{3}{*}{ UF  }
                         & Fair CATE        &  0.046 (0.034)   & 0.037 (0.028)   & 0.035 (0.024) \\
                           & AF-IDR      &0.030   (0.026) &0.028   (0.022) & 0.027   (0.020)  \\
                         & DP-IDR   &0.046   (0.037)   &0.039   (0.029)  &   0.034   (0.027)  \\
 \cmidrule{1-5}
  \multirow{3}{*}{ PV }
                         & Fair CATE   &1.432 (0.065)     &
                    1.431 (0.064)  &1.430 (0.058)    \\
     & AF-IDR   &      1.516   (0.266)         & 1.615   (0.164) &1.625   (0.160)  \\
                         & DP-IDR  &  1.875   (0.079)  & 1.887   (0.074) &  1.911   (0.069) \\
\bottomrule
\end{tabular}
\end{table*}
\begin{table*}[h]
\caption{ Sample means and sample standard deviations (in parentheses) of the unfairness levels (UF) and policy value (PV) for the proposed DP-IDR, Fair CATE and AF-IDR methods, based on 200 replications under Case 2.}
  \centering
\begin{tabular}{clccc}
   \toprule
 & Method  & n=500 & n=1000 & n=2000 \\
       \hline

              \hline

 \multirow{3}{*}{ UF }
                        & Fair CATE      & 0.129 (0.091)  &0.124 (0.068)  &  0.121 (0.054)  \\
                              & AF-IDR    & 0.046   (0.032)  &0.045   (0.031)  &0.040   (0.030)
                              \\
                        & DP-IDR & 0.052   (0.033) &0.036   (0.029)  & 0.033   (0.027) \\

 \cmidrule{1-5}

   \multirow{3}{*}{PV}
                        & Fair CATE   &0.616 (0.115)   &0.639 (0.107)  &  0.652 (0.100)  \\
                            &  AF-IDR  & 0.479   (0.180) & 0.528   (0.128) & 0.558   (0.119)  \\
                       & DP-IDR   &0.646   (0.121)  &0.705   (0.103) & 0.742   (0.099)\\
\bottomrule
\end{tabular}
\end{table*}

\begin{table*}[h]
\caption{
Sample means and sample standard deviations (in parentheses) of the unfairness levels (UF) and policy value (PV) for the proposed $\epsilon$-DP-IDR method based on 200 replications when $n=2000$ under Cases 1 and  2.}
\centering
\begin{tabular}{ccccccc}
\toprule
  &   \multicolumn{2}{c}{ Case 1}  & \multicolumn{2}{c}{Case 2 } \\
  \cmidrule(lr){2-3} \cmidrule(lr){4-5}
%\hline
 $\epsilon$ & UF & PV &  UF &  PV  \\
 \cmidrule(lr){1-3} \cmidrule(lr){4-5}

  0.02 &  0.038   (0.030) &  1.919   (0.082)  &0.041   (0.030)& 0.754   (0.099) \\
 0.04 & 0.049   (0.035)    &1.925   (0.082)
  & 0.051   (0.036)  & 0.764   (0.099)\\
 0.08 &  0.081   (0.041) &1.937   (0.081)
& 0.082   (0.043) & 0.786   (0.100)\\
 0.10 & 0.100   (0.041)    &  1.941   (0.080)   & 0.104   (0.042) &0.795   (0.099) \\
 0.15 & 0.147   (0.041)   & 1.946   (0.079)  & 0.149   (0.041) &0.818   (0.099) \\
\bottomrule
\end{tabular}
\end{table*}
\begin{table*}[h]
\caption{
Sample means and sample standard deviations (in parentheses) of the conditional unfairness levels (CUF) and policy value (PV) for the proposed CDP-IDR method based on 200 replications  under Cases 3 and 4. }
  \centering
\begin{tabular}{cccccc}
   \toprule
 &            &n=500           & n=1000  &   n=2000 \\
       \hline
 \multirow{2}{*}{  Case 3} & CUF
   & 0.054   (0.032) & 0.043   (0.025)  & 0.042   (0.022)            \\

 & PV &0.022   (0.060) &  0.124 (0.054)  & 0.199   (0.045)

                                     \\
                                            \hline
\multirow{2}{*}{  Case 4}
& CUF
  & 0.074   (0.051) & 0.068   (0.045)&  0.049   (0.026)
             \\

 & PV
        & 1.668   (0.129) &  1.789   (0.121)  & 1.934   (0.096)   \\        \bottomrule

\end{tabular}
\end{table*}
\begin{table*}[h]
\caption{Sample means and sample standard deviations (in parentheses) of the conditional unfairness levels (CUF) and policy value (PV) for the proposed $\epsilon$-CDP-IDR method based on 200 replications when $n=2000$  under Cases 3 and 4.}
\centering
\begin{tabular}{ccccc}
\toprule

   &   \multicolumn{2}{c}{ Case 3} & \multicolumn{2}{c}{Case 4 } \\
  \cmidrule(lr){2-3} \cmidrule(lr){4-5}
$\epsilon$  & CUF  &PV  &CUF  &  PV\\
  \cmidrule(lr){1-3} \cmidrule(lr){4-5}

0.05 &  0.047   (0.022)   & 0.228   (0.045)  &0.063   (0.026) &1.979   (0.094) \\
0.10 & 0.083   (0.026)     &  0.253  (0.043) &  0.102   (0.032)  &2.015   (0.093)\\
0.15 &  0.128   (0.029)          &  0.271  (0.046) &0.150   (0.031) &2.045   (0.091)\\
0.20 &  0.163   (0.030)  &    0.285   (0.045) &0.200   (0.031) & 2.067   (0.089)\\
0.25 & 0.191  (0.031)        & 0.295   (0.045) &0.246   (0.031) & 2.083   (0.087) \\
\bottomrule
\end{tabular}
\end{table*}

\subsubsection{Performance Evaluation of DP-IDR and $\epsilon$-DP-IDR Estimations}
Tables 1 and 2 show that the proposed DP-IDR method consistently achieves the highest policy value across all sample sizes under Cases 1 and 2, outperforming both the Fair CATE and AF-IDR methods. The empirical results align with the theoretical intuition that directly constraining the IDR—rather than the CATE—imposes fewer restrictions and retains greater flexibility in estimating the IDR. The Fair CATE method enforces demographic parity for CATE, which may overly constrain the corresponding IDR. The AF-IDR method, which relies on the learned representation satisfying demographic parity, yields the lowest policy value in Case 2. In reducing unfairness levels, the DP-IDR method demonstrates competitive performance compared to alternative approaches.

Table 3 presents the performance of the proposed $\epsilon$-DP-IDR method under Cases 1 and 2. As $\epsilon$ decreases, the unfairness level also decreases and is effectively controlled below or near the specified threshold.

\subsubsection{Performance Evaluation of CDP-IDR and $\epsilon$-CDP-IDR Estimations}
Table 4 evaluates the performance of the CDP-IDR method under Cases 3 and 4. Across all sample sizes, the CDP-IDR method attains   conditional
unfairness level below  $0.05$  when $n=2000$, demonstrating its effectiveness in reducing unfairness given legitimate covariates.

Table 5 presents the performance of the $\epsilon$-CDP-IDR method. The conditional unfairness level remains  below or near the pre-specified threshold $\epsilon$, demonstrating the method’s ability to control conditional unfairness level.

\section{Application}
We used the proposed methods to analyze the real data from the Oregon Health Insurance Experiment (OHIE) \citep{finkelstein2012oregon}. The dataset can be downloaded at the web link: {https://www.nber.org/research/data/oregon-health-insurance-experiment-data}.
In 2008, the state of Oregon offered Medicaid to low-income and uninsured adults through a lottery.  Individuals selected by the lottery  had the opportunity to apply for Medicaid. Eligible participants  were provided with free healthcare, including physician services, prescription drugs, mental health, and chemical dependency services. After 12 months, a mail survey was conducted to assess various outcomes including financial strain.

In our study,
the treatment $A=1$ indicated the receipt of free healthcare through the Medicaid program, and the outcome $R$ was the negative value of the total amount owed for medical expenses. When individuals provided materials in English, set $S=1$; otherwise, set $S=0$. The covariates $\boldsymbol{X}$ included individual-level medical history variables, such as the number of emergency visits prior to the experiment, overall health level, and whether the individual had any basic diseases. The covariates also included demographic variables, such as age, income, family size, education level, and language. The dimension of covariates is $24$. The group variable \(L = 1, 2 \text{ or } 3 \)  indicates whether an individual's income is less than \(10\%\), between \(10\%\) and  \(20\%\), or greater than  \(20\%\) of the federal poverty line, respectively.
After removing  samples with missing values,  we randomly split the $7135$  observations into training ($72\%$), validation ($8\%$), and testing $(20\%)$ sets. For any estimated decision function $\widehat{f}$, its policy value  was estimated by  $ \sum_j\left\{\widehat{\delta}_R(\boldsymbol{Z}_j)I(\widehat{f}(\boldsymbol{Z}_j)>0)\right\}/N$, where  $\widehat{\delta}_R$ was estimated using a  deep neural network and  $\boldsymbol{Z}_j, j=1, 2, \ldots, N,$ were samples from the testing set.

\begin{table*}[h]
\caption{ Sample means and sample standard deviations (in parentheses) of
the unfairness levels (UF) and the policy value (PV) for the proposed $\epsilon$-DP-IDR and Fair CATE approaches, based on 5-fold cross-validation in the OHIE dataset.}
  \centering
\begin{tabular}{lccc}
\toprule
 &$\epsilon$ &  { UF} & {PV} \\
\hline
Fair CATE & -- & 0.047 (0.041) & 707.934 (74.358) \\
AF-IDR & -- & 0.020   (0.033)& 961.090   (38.755) \\
\midrule
\multirow{9}{*}{$\epsilon$-DP-IDR }
&$0$ &   {0.017}  (0.025) &  {972.877}   (43.622) \\
&$0.05$ & 0.068   (0.065) &976.658   (44.637) \\
&$0.10$ & 0.114   (0.072) & 977.907   (45.760)\\
&$0.15$ & 0.162   (0.069) & 980.630   (44.854)\\
&$0.20$ & 0.192   (0.069) & 981.555   (43.818)\\
&$0.25$ & 0.234   (0.068)& 982.233   (43.424) \\
&$0.30$ &0.271   (0.080)& 981.849   (44.103)\\
\bottomrule
\end{tabular}
\end{table*}
Table 6 reports the unfairness levels and policy values for DP-IDR and $\epsilon$-DP-IDR methods. The proposed DP-IDR method achieves the lowest unfairness level and the highest policy value, outperforming both Fair CATE and AF-IDR methods. In contrast, Fair CATE and AF-IDR methods incur a substantial reduction in policy values.
Moreover, we observe a trade-off governed by $\epsilon$, as $\epsilon$ decreases, the unfairness level is maintained below or near $\epsilon$, while the policy value correspondingly decreases slightly.
\begin{table*}[h]
\caption{Sample means and sample standard deviations (in parentheses) of
the conditional unfairness levels (CUF) and the policy value (PV) for the proposed $\epsilon$-CDP-IDR method, based on 5-fold cross-validation in the OHIE dataset.}
  \centering
\begin{tabular}{cccc}
\toprule
  & $\epsilon$  & CUF & PV \\
\hline
  \multirow{9}{*}{$\epsilon$-CDP-IDR }
 & $0$  & 0.066 (0.086) & 987.641   (57.555) \\
&$0.05$ & 0.074 (0.084) &988.958   (59.573) \\
&$0.10$ & 0.117 (0.105) & 990.738   (60.975) \\
&$0.15$ & 0.148 (0.138) &991.778   (61.837)\\
&$0.20$ &0.218   (0.207) & 991.930   (61.757)\\
&$0.25$ &0.274   (0.246) & 992.802   (62.186) \\
&$0.30$ &  0.288   (0.257) &  992.642   (61.818) \\
\bottomrule
\end{tabular}
\end{table*}

Table 7 presents the results for CDP-IDR  and $\epsilon$-CDP-IDR methods, which aims to reduce  the conditional unfairness level.  When $\epsilon = 0$, the CDP-IDR method achieves a substantial reduction in  the conditional unfairness level  while preserving a high policy value.  The $\epsilon$-CDP-IDR method effectively controls the unfairness level below or near the specified unfairness level $\epsilon$.

\section{Discussion}
Fairness is a critical consideration when implementing individualized decision rules in our society. In this work, we propose methods that incorporate demographic parity  and conditional demographic parity into the estimation of IDRs. Our approach is conceptually equivalent to applying fairness-aware perturbations to the optimal IDRs, thereby ensuring fairness constraints are met during policy learning.
To the best of our knowledge, this is the first framework that explicitly integrates demographic parity directly into the IDR optimization process.

Three directions for future work are of particular interest. First, while our current formulation assumes binary sensitive attributes, extending  DP-ITR or CDP-IDR to accommodate categorical or continuous sensitive variables would significantly broaden its applicability.
 Second, the proposed framework can be  extended to incorporate  other fairness constraints. One example  is value fairness \citep{GITR} constraint, which requires that the average benefit (or risk) of the implemented IDR is similar across subgroups. Let the observed data be $\{\boldsymbol{X}, S, A, R\}$, and
define the policy value in the  $s$-th group  as $E\left[I(f(\boldsymbol{X},S)>0)\delta_R(\boldsymbol{X},S)|S=s \right]$. For $s=1$,  the subgroup policy value can be expressed as
\begin{equation}
\begin{aligned}
\label{e.9}
E\left[I(f(\boldsymbol{X},S)>0)\delta_R(\boldsymbol{X},S)|S=1 \right]
=&E\left[I(f(\boldsymbol{X},S)>0)\delta_R(\boldsymbol{X},S)\frac{I(S=1)}{P(S=1)}\right].
\nonumber
\end{aligned}
\end{equation}
Under value fairness constraint, the optimal IDR solves
\begin{equation}
\begin{aligned}
\label{value_fairness}
 \max_f  & E\left[I(f(\boldsymbol{X},S)>0)\delta_R(\boldsymbol{X},S) \right]\\
 \text{ subject to }
&  E\left[I(f(\boldsymbol{X}, S)>0) \delta_R(\boldsymbol{X},S)\left\{\frac{I(S=1)}{P(S=1)}-\frac{I(S=0)}{P(S=0)} \right\} \right]  =0.
\end{aligned}
\end{equation}
Following the derivations in Section 3.1,
the solution $f_{0}$ to Equation \eqref{value_fairness} satisfies $$I[f_{0}(\boldsymbol{X},S)>0]=I\left[\delta_R(\boldsymbol{X},S)\left\{1- \lambda^{*} \left(\frac{I(S=1)}{P(S=1)}-\frac{I(S=0)}{P(S=0)} \right) \right\}>0   \right],$$
 where $\lambda^{*}$ solves  {\small
$$E\left[I\left\{\delta_R(\boldsymbol{X},S)\left\{1- \lambda \left(\frac{I(S=1)}{P(S=1)}-\frac{I(S=0)}{P(S=0)} \right) \right\}>0   \right\}\delta_R(\boldsymbol{X},S)\left\{\frac{I(S=1)}{P(S=1)}-\frac{I(S=0)}{P(S=0)} \right\} \right]=0.$$}
Thus, the optimal IDR that satisfies value fairness constraints  admits the closed-form expression $2I(f_0(\boldsymbol{X}, S)>0)-1$.
Finally, fairness-constrained optimization problems may be formulated  within the outcome-weighted learning \citep{zhao2012estimating} or A-learning \citep{murphy2003optimal} frameworks, where deep neural networks can be employed to accommodate high-dimensional and nonlinear data structures.

\bibliographystyle{chicago}
\bibliography{Bibliography-MM-MC}

\end{document}